\documentclass[final]{cvpr}

\usepackage{times}
\usepackage{epsfig}
\usepackage{graphicx}
\usepackage{amsmath}
\usepackage{amssymb}
\usepackage{ctable}
\usepackage{color,colortbl}
\usepackage[ruled,vlined,linesnumbered]{algorithm2e}
\usepackage[symbol]{footmisc}

\definecolor{Gray}{gray}{0.9}
\definecolor{White}{gray}{1}

\usepackage[breaklinks=true,bookmarks=false]{hyperref}

\usepackage{xcolor}

\def\cvprPaperID{1017} 
\def\confYear{CVPR 2021}

\begin{document}

\title{HybrIK: A Hybrid Analytical-Neural Inverse Kinematics Solution \\for 3D Human Pose and Shape Estimation}

\author{{Jiefeng Li},~ {Chao Xu},~ {Zhicun Chen},~ {Siyuan Bian},~ {Lixin Yang},~ {Cewu Lu}\footnotemark[2]\\
{Shanghai Jiao Tong University, China}\\
{\tt\small \{{ljf\_likit},{xuchao.19962007}, {zhicun\_chen}, {biansiyuan}, {siriusyang}, {lucewu}\}@{sjtu.edu.cn}}
}

\maketitle


\begin{abstract}
   Model-based 3D pose and shape estimation methods reconstruct a full 3D mesh for the human body by estimating several parameters. However, learning the abstract parameters is a highly non-linear process and suffers from image-model misalignment, leading to mediocre model performance. In contrast, 3D keypoint estimation methods combine deep CNN network with the volumetric representation to achieve pixel-level localization accuracy but may predict unrealistic body structure. In this paper, we address the above issues by bridging the gap between body mesh estimation and 3D keypoint estimation. We propose a novel hybrid inverse kinematics solution (HybrIK). HybrIK directly transforms accurate 3D joints to relative body-part rotations for 3D body mesh reconstruction, via the twist-and-swing decomposition. The swing rotation is analytically solved with 3D joints, and the twist rotation is derived from the visual cues through the neural network. We show that HybrIK preserves both the accuracy of 3D pose and the realistic body structure of the parametric human model, leading to a pixel-aligned 3D body mesh and a more accurate 3D pose than the pure 3D keypoint estimation methods. Without bells and whistles, the proposed method surpasses the state-of-the-art methods by a large margin on various 3D human pose and shape benchmarks. As an illustrative example, HybrIK outperforms all the previous methods by \textbf{13.2} mm MPJPE and \textbf{21.9} mm PVE on 3DPW dataset. Our code is available at \href{https://github.com/Jeff-sjtu/HybrIK}{https://github.com/Jeff-sjtu/HybrIK}.
\end{abstract}

\footnotetext[2]{Cewu Lu is the corresponding author. He is the member of Qing Yuan Research Institute, Qi Zhi Institute and MoE Key Lab of Artificial Intelligence, AI Institute, Shanghai Jiao Tong University, China.}


\section{Introduction}

Recovering the 3D surface from a monocular RGB image is a fundamentally ill-posed problem. It has a wide rage of application scenarios~\cite{activity0,activity4,li2020detailed,li2020pastanet,avatar0}. With the development of the parametric statistical human body shape models~\cite{anguelov2005scape,loper2015smpl,pavlakos2019expressive}, a realistic and controllable 3D mesh of human body can be generated from only a few parameters, \eg shape parameters and relative rotations of body parts. Recent studies develop the model-based methods~\cite{bogo2016keep,hmr,spin} to obtain these parameters from the monocular RGB input and produce 3D pose and shape of human bodies.

\begin{figure}[t]
    \begin{center}
        \includegraphics[width=\linewidth]{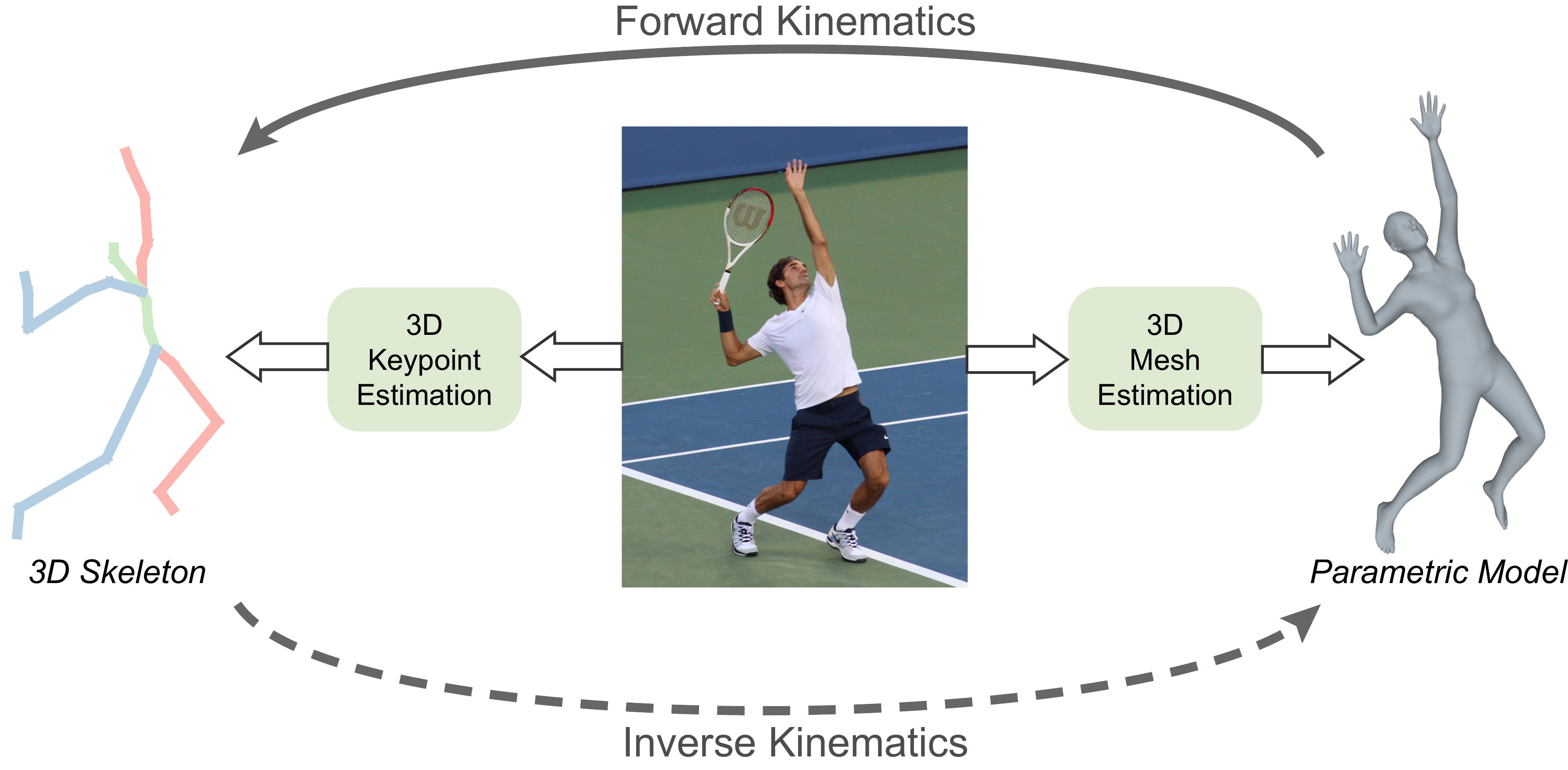}
    \end{center}
    \vspace{-3mm}
    \caption{\textbf{Closing the loop between the 3D skeleton and the parametric model via HybrIK.} 
    A 3D skeleton predicted by the neural network can be transformed into a parametric body mesh by inverse kinematics without loss of accuracy. The parametric body mesh can generate structural realistic 3D skeleton by forward kinematics.}
    \label{fig:loop}
    \vspace{-3mm}
\end{figure}

Most of the model-based methods can be categorized into two classes: optimization-based approach and learning-based approach. Optimization-based approaches~\cite{guan2009estimating,bogo2016keep,pavlakos2019expressive} estimate the body pose and shape by an iterative fitting process. The parameters of the statistical model are tuned to reduce the error between its 2D projection and 2D observations, \eg 2D joint locations and silhouette.
However, the optimization problem is non-convex and takes a long time to solve. The results are sensitive to the initialization. These issues shift the spotlight towards the learning-based approaches.
With a parametric body model, learning-based approaches use neural networks to regress the model parameters directly~\cite{hmr,spin,vibe}. But the parameter space in the statistical model is abstract, making it difficult for the networks to learn the mapping function.


This challenge prompts us to look into the field of 3D keypoint estimation. Instead of the direct regression, previous methods~\cite{coarse,integral} adopt volumetric heatmap as the target representation to learn 3D joint locations and have achieved impressive performance. This inspires us to build a collaboration between the 3D joints and the body mesh (Fig.~\ref{fig:loop}). On the one hand, the accurate 3D joints facilitate the 3D body mesh estimation. On the other hand, the shape prior in parametric body model in turn fixes the unrealistic body structure issue of the 3D keypoint estimation methods.
Since the current 3D keypoint estimation methods lack explicit modelling of the distribution of body bone length, it may predict unrealistic body structures like left-right asymmetry and abnormal proportions of limbs.
By leveraging the parametric body model, the presented human shape better conforms to the actual human body.

In this work, we propose a hybrid analytical-neural inverse kinematics solution (HybrIK) to bridge the gap between 3D keypoint estimation and body mesh estimation. Inverse kinematics (IK) process is the mathematical process of finding the relative rotations to produce the desired locations of body joints. It is an ill-posed problem because there is no unique solution.
The core of our approach is to propose an innovative IK solution via twist-and-swing decomposition.
The relative rotation of a skeleton part is decomposed into \textit{twist} and \textit{swing}, \ie a longitudinal rotation and an in-plane rotation. In HybrIK, we composite the entire rotation recursively along the kinematic tree by analytically calculating \textit{swing} rotation and predicting \textit{twist} rotation.
A critical characteristic of our approach is that the relative rotation estimated by HybrIK is naturally aligned with the 3D skeleton, without the need for additional optimization procedures in the previous approaches~\cite{bogo2016keep,pavlakos2019expressive,spin}.
All operations in HybrIK are differentiable, which allows us to simultaneously train 3D joints and human body mesh in an end-to-end manner.
Besides, experiments indicate that HybrIK raises the performance of body mesh estimation to the same level as 3D keypoint estimation and takes a step forward. The proposed approach is benchmarked in various 3D human pose and shape datasets, and it significantly outperforms state-of-the-art approaches~\cite{spin,moon2020i2l} by \textbf{21.9} mm PVE on 3DPW~\cite{3dpw}, \textbf{6.6} mm PA-MPJPE on Human3.6M~\cite{h36m} and \textbf{10.8} AUC on MPI-INF-3DHP~\cite{3dhp}.

The contributions of our approach can be summarized as follows:
\begin{itemize}
    \item We propose HybrIK, a hybrid analytical-neural IK solution that converts the accurate 3D joint locations to full 3D human body mesh. HybrIK is differentiable and allows end-to-end training.
    \item Our approach closes the loop between the 3D skeleton and the parametric model. It fixes the alignment issue of current model-based body mesh estimation methods and the unrealistic body structure problem of 3D keypoint estimation methods at the same time.
    \item Our approach achieves state-of-the-art performance across various 3D human pose and shape benchmarks.
\end{itemize}

\section{Related Work}

\paragraph{3D Keypoint Estimation.}
Many works formulate 3D human pose estimation as the problem of locating the 3D joints of the human body.
Previous studies can be divided into two categories: single-stage and two-stage approaches.
Single-stage approaches~\cite{pavlakos2017coarse,rogez2017lcr,mehta2017monocular,zhou2017towards,mehta2018single,sun2018integral,moon2019camera,wang2020hmor} directly estimate the 3D joint locations from the input image.
Various representations are developed, including 3D heatmap~\cite{pavlakos2017coarse}, location-map~\cite{mehta2017monocular} and 2D heatmap + $z$ regression~\cite{zhou2017towards}.
Two-stage approaches first estimate 2D pose and then lift them to 3D joint locations by a learned dictionary of 3D skeleton~\cite{akhter2015pose,ramakrishna2012reconstructing,tung2017adversarial,sanzari2016bayesian,zhou2016sparse,zhou2016sparseness} or regression~\cite{park20163d,yasin2016dual,moreno20173d,fang2017learning,martinez2017simple,sun2017compositional}. Two-stage approaches highly rely on the accurate 2D pose estimators, which have achieved impressive performance by the combination of powerful backbone network~\cite{vgg,resnet,hourglass,pang2019deep,pang2020complex} and the 2D heatmap.


These privileged forms of supervision contribute to the recent performance leaps of 3D keypoint estimation. However, the human structural information is modelled implicitly by the neural network, which can not ensure the output 3D skeletons to be realistic. Our approach combines the advantages of both the 3D skeleton and parametric model to predict accurate and realistic human pose and shape.

\vspace{-3mm}
\paragraph{Model-based 3D Pose and Shape Estimation.} Pioneer works on the model-based 3D pose and shape estimation methods use parametric human body model~\cite{anguelov2005scape,loper2015smpl,pavlakos2019expressive} as the output target because they capture the statistics prior of body shape. Compared with the model-free methods~\cite{varol2018bodynet,kolotouros2019convolutional,moon2020i2l}, the model-based methods directly predict controllable body mesh, which can facilitate many downstream tasks for both computer graphics and computer vision. Bogo \etal~\cite{bogo2016keep} propose SMPLify, a fully automatic approach, without manual user intervention~\cite{sigal2008combined,guan2009estimating}. This optimization paradigm was further extended with silhouette cues~\cite{lassner2017unite}, volumetric grid~\cite{varol2018bodynet}, multiple people~\cite{zanfir2018monocular} and whole-body parametric model~\cite{pavlakos2019expressive}.

With the advances of the deep learning networks, there are increasing studies that focus on the learning-based methods, using a deep network to estimate the pose and shape parameters. Since the mapping from RGB image to shape space and relative body-part rotation is hard to learn, many works use some form of intermediate representation to alleviate this problem, such as keypoints and silhouettes~\cite{pavlakos2018learning}, semantic part segmentation~\cite{omran2018neural} and 2D heatmap input~\cite{tung2017self}. Kanazawa \etal~\cite{hmr} use an adversarial prior and an iterative error feedback (IEF) loop to reduce the difficulty of regression. Arnab \etal~\cite{arnab2019exploiting} and Kocabas \etal~\cite{vibe} exploit temporal context, while Guler \etal~\cite{guler2019holopose} use a part-voting expression and test-time post-processing to improve the regression network. Kolotouros \etal~\cite{spin} leverage the optimization paradigm to provide extra 3D supervision from unlabeled images. 

In this work, we address this challenging learning problem by a transformation from the pixel-aligned 3D joints to the relative body-part rotations.

\vspace{-5mm}
\paragraph{Body-part Rotation in Pose Estimation.} The core of our approach is to calculate the relative rotation of human body parts through a hybrid IK process. There are several works that estimate the relative rotations in the 3D pose estimation literature. Zhou \etal~\cite{zhou2016deep} use the network to predict the rotation angle of each body joint, followed by an FK layer to generate the 3D joint coordinates. Pavllo \etal~\cite{pavllo2018quaternet} switch to quaternions, while Yoshiyasu \etal~\cite{yoshiyasu2018skeleton} directly predict the $3\times3$ rotation matrices. Mehta \etal~\cite{mehta2019xnect} first estimate the 3D joints and then use a fitting procedure to find the rotation Euler angles.
Previous approaches are either limited to a hard-to-learn problem or require an additional fitting procedure. Our approach recovers the body-part rotation from 3D joint locations in a direct, accurate and feed-forward manner.

\vspace{-5mm}
\paragraph{Inverse Kinematics Process.} The inverse kinematics (IK) problem has been extensively studied during recent decades. Numerical solutions~\cite{balestrino1984robust,wolovich1984computational,girard1985computational,klein1983review,wampler1986manipulator,buss2005selectively} are simple ways to implement the IK process, but they suffer from time-consuming iterative optimization.
Heuristic methods are efficient solutions to the IK problem. For example, CDC\cite{luenberger1984linear}, FABRIK\cite{aristidou2011fabrik} and IK-FA\cite{rokbani2015ik} have a low computational cost for each heuristic iteration.
In some special cases, there exist analytical solutions to the IK problem. Tolani \etal~\cite{tolani2000real} propose a reliable algorithm by the combination of analytical and numerical methods. Kallmann \etal~\cite{kallmann2008analytical} solve the IK for arm linkage, \ie a three-joint system. Recently, researchers have been interested in using neural networks to solve the IK problem in robotic control~\cite{csiszar2017solving}, motion retargeting~\cite{villegas2018neural} and hand pose estimation~\cite{mueller2017real,kokic2019learning}.

In this work, we combine the interpretable characteristic of analytical solution and the flexibility of the neural network, introducing a feed-forward hybrid IK algorithm with twist-and-swing decomposition. Twist-and-swing decomposition is introduced by Baerlocher \etal~\cite{baerlocher2001parametrization}. The twist angles are limited based on the particular body joint. In our works, the twist angles are estimated by neural networks, which is more flexible and can be generalized to all body joints. Compared with previous analytical solutions~\cite{kallmann2008analytical} designed for specific joint linkage, our algorithm can be applied to the entire body skeleton in a direct and differentiable manner.

\section{Method}

\begin{figure*}[t]
    \begin{center}
        \includegraphics[width=0.95\linewidth]{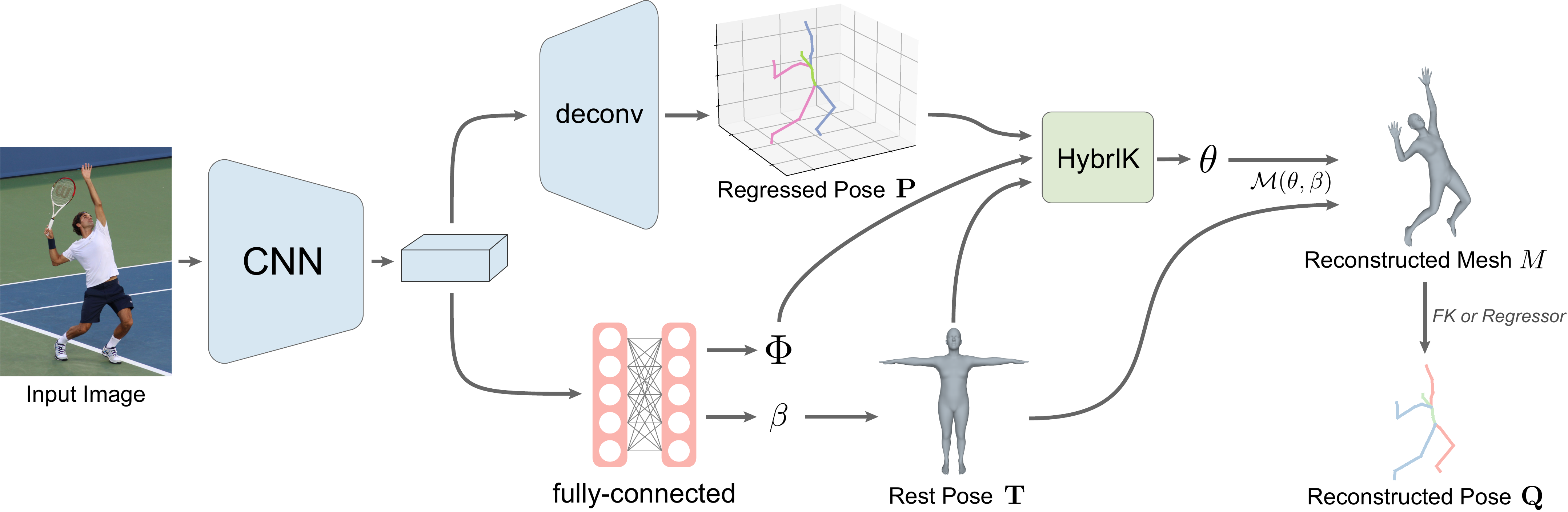}
    \end{center}
    \vspace{-3mm}
    \caption{\textbf{Overview of the proposed framework.} A 3D heatmap is generated by the deconvolution layers and used to regress the 3D joints $\mathbf{P}$. The shape parameters $\beta$ and the \textit{twist} angle $\Phi$ are learned from the visual cues through the fully-connected layers. These results are then sent to the HybrIK process to solve the relative rotation, \ie the pose parameters $\theta$. Finally, with the pose and shape parameters, we can obtain the reconstructed body mesh $M$, and the reconstructed pose ${\mathbf{Q}}$ via a further FK process or linear regression.
    }
    \label{fig:pipeline}
    \vspace{-3mm}
\end{figure*}

In this section, we present our hybrid analytical-neural inverse kinematics solution that boosts 3D human pose and shape estimation (Fig.~\ref{fig:pipeline}). First, in \S\ref{sec:bg}, we briefly describe the forward kinematics process, the inverse kinematics process and the SMPL model. In \S\ref{sec:pat-ik}, we introduce the proposed inverse kinematics solution, HybrIK. Then, in \S\ref{sec:learning}, we present the overall learning framework to estimate the pixel-aligned body mesh and realistic 3D skeleton. Finally, we provide the necessary implementation details in \S\ref{sec:implement}.

\subsection{Preliminary}\label{sec:bg}

\paragraph{Forward Kinematics.} Forward kinematics (FK) for human pose usually refers to the process of computing the reconstructed pose $\mathbf{Q} = \{{q}_k\}_{k=1}^{K}$, with the rest pose template $\mathbf{T} = \{{t}_k\}_{k=1}^{K}$ and the relative rotations $\mathbf{R} = \{R_{{\mathtt{pa}(k)},{k}}\}_{k=1}^{K}$ as input:
\begin{equation}
    \mathbf{Q} = \text{FK}(\mathbf{R}, \mathbf{T}),
    \label{eq:fk}
\end{equation}
where $K$ is the number the body joints, $q_k \in \mathbb{R}^3$ denotes the reconstructed 3D location of the $k$-th joint, $t_k \in \mathbb{R}^3$ denotes the $k$-th joint location of the rest pose template, $\mathtt{pa}(k)$ return the parent's index of the $k$-th joint, and $R_{{\mathtt{pa}(k)},{k}}$ is the relative rotation of $k$-th joint with respect to its parent joint. FK can be performed by recursively rotating the template body part from the root joint to the leaf joints:
\begin{equation}
    q_k = R_k(t_{k} - t_{\mathtt{pa}(k)}) + q_{\mathtt{pa}(k)},
    \label{eq:fk-step}
\end{equation}
where $R_{k} \in \mathbb{SO}(3)$ is the global rotation of the $k$-th joint with respect to the canonical rest pose space.
The global rotation can be calculated recursively:
\begin{equation}
    R_k = R_{\mathtt{pa}(k)} R_{\mathtt{pa}(k), k}.
    \label{eq:r-step}
\end{equation}
For the root joint that has no parent, we have $q_0 = t_0$.

\paragraph{Inverse Kinematics.} Inverse kinematics (IK) is the reverse process of FK, computing relative rotations $\mathbf{R}$ that can generate the desired locations of input body joints $\mathbf{P} = \{p_k\}_{k=1}^{K}$. This process can be formulated as:
\begin{equation}
    \mathbf{R} = \text{IK}(\mathbf{P}, \mathbf{T}),
\end{equation}
where $p_k$ denotes the $k$-th joint of the input pose. Ideally, the resulting rotations should satisfy the following condition:
\begin{equation}
    p_k - p_{\mathtt{pa}(k)} = R_k(t_k - t_{\mathtt{pa}(k)}) \quad \forall 1 \leq k \leq K.
    \label{eq:ik-cond}
\end{equation}
Similar to the FK process, we have $p_0 = t_0$ for the root joint that has no parent. While the FK problem is well-posed, the IK problem is ill-posed because there is either no solution or because there are many solutions to fulfill the target joint locations.

\paragraph{SMPL Model.} In this work, we employ the SMPL~\cite{loper2015smpl} parametric model for human body representation. SMPL allows us to use shape parameters and pose parameters to control the full human body mesh. The shape parameters $\beta \in \mathbb{R}^{10}$ are parameterized by the first $10$ principal components of the shape space. The pose parameters $\theta$ are modelled by relative 3D rotation of $K = 23$ joints, $\theta = (\theta_1,\theta_2,\cdots,\theta_K)$.
SMPL provides a differentiable function $\mathcal{M}(\theta, \beta)$ that takes the pose parameters $\theta$ and the shape parameters $\beta$ as input and outputs a triangulated mesh $M \in \mathbb{R}^{N \times 3}$ with $N=6980$ vertices. Conveniently, the reconstructed 3D joints ${\mathbf{Q}}_{\textit{smpl}}$ can be obtained by an FK process, \ie ${\mathbf{Q}}_{\textit{smpl}} = \text{FK}(\mathbf{R}, \mathbf{T})$. Also, the joints of Human3.6M~\cite{h36m} can be obtained by a linear combination of the mesh vertices through a linear regressor $W$, \ie ${\mathbf{Q}}_{\textit{h36m}} = WM$.

\subsection{Hybrid Analytical-Neural Inverse Kinematics}\label{sec:pat-ik}
Estimating the human body mesh by direct regression of the relative rotations is too difficult~\cite{hmr,spin,vibe}. Here, we propose a hybrid analytical-neural inverse kinematics solution (HybrIK) to leverage 3D keypoints estimation to boost 3D body mesh estimation.
Since 3D joints cannot uniquely determine the relative rotation, we decompose the original rotation into \textit{twist} and \textit{swing}. The 3D joints are utilized to calculate the \textit{swing} rotation analytically, and we exploit the visual cues by a neural network to estimate the 1-DoF \textit{twist} rotation. In HybrIK, the relative rotations are solved recursively along the kinematic tree. We conduct error analysis and further develop an adaptive solution to reduce the reconstruction error.


\begin{figure}[tb]
    \begin{center}
        \includegraphics[width=0.95\linewidth]{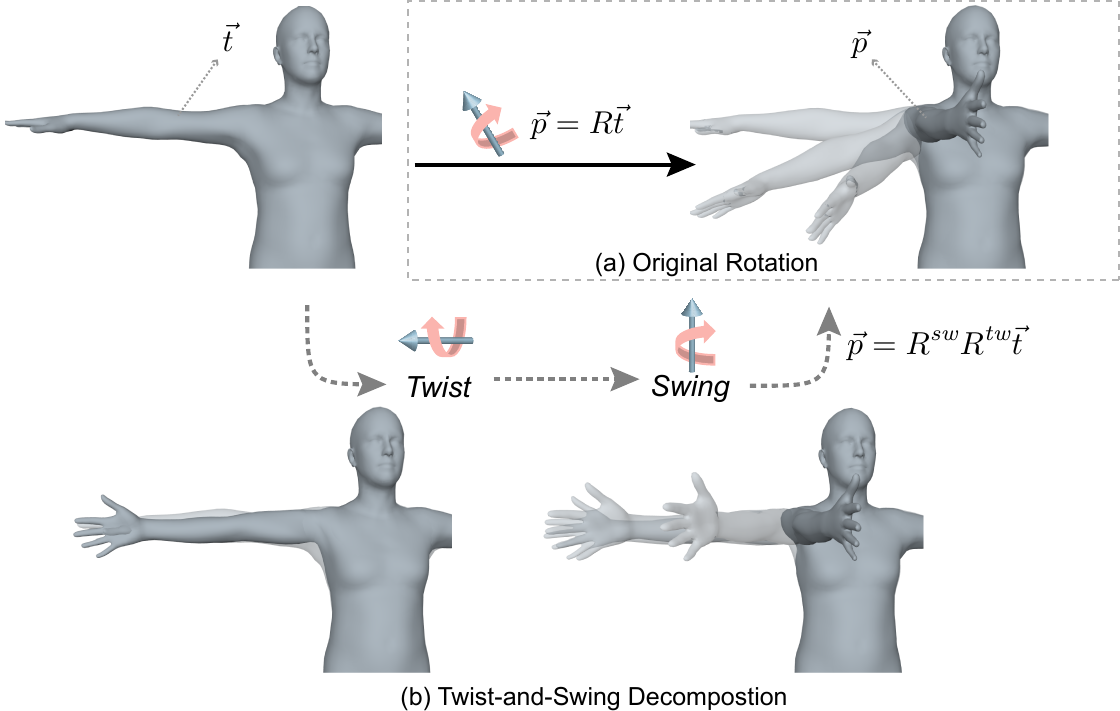}
    \end{center}
    \vspace{-3mm}
    \caption{\textbf{Illustration of the twist-and-swing decomposition.} (a) The original rotation turns the right palm-down hand to the front and the palm to the left in one step. (b) With twist-and-swing decomposition, the rotation can be divided into two steps: First, turn the palm $90^\circ$, and then move the entire hand to the front.}
    \label{fig:pat}
    \vspace{-3mm}
\end{figure}

\vspace{-2mm}
\paragraph{Twist-and-Swing Decomposition.}
In the analytical IK formulation, some body joints are usually assigned lower degree-of-freedom (DoFs) to simplify the problem, \eg $1$ or $2$ DoFs~\cite{kofinas2013complete,tolani2000real,kallmann2008analytical}. In this work, we consider a general case where each body joint is assumed to have full $3$ DoFs. As illustrated in Fig.~\ref{fig:pat}, a rotation $R \in \mathbb{SO}(3)$ can be decomposed into a \textit{twist} rotation $R^{\textit{tw}}$ and a \textit{swing} rotation $R^{\textit{sw}}$. 
Given the start template body-part vector $\vec{t}$ and the target vector $\vec{p}$, the solution process of $R$ can be formulated as:
\begin{equation}
    R = \mathcal{D}(\vec{p}, \vec{t}, \phi) = \mathcal{D}^{\textit{sw}}(\vec{p}, \vec{t})\mathcal{D}^{\textit{tw}}(\vec{t}, \phi) = R^{\textit{sw}}R^{\textit{tw}},
    \label{eq:dec}
\end{equation}
where $\phi$ is the \textit{twist} angle that estimated by a neural network, $\mathcal{D}^{\textit{sw}}(\cdot)$ is a closed-form solution of the \textit{swing} rotation, and $\mathcal{D}^{\textit{tw}}(\cdot)$ transforms $\phi$ to the \textit{twist} rotation. Here, $R$ should satisfy the condition in Eq.~\ref{eq:ik-cond}, \ie $\vec{p} = R \vec{t}$.

\textit{\textbf{- Swing}:}\quad
The \textit{swing} rotation has the axis $\vec{n}$ that is perpendicular to $\vec{t}$ and $\vec{p}$. Therefore, it can be formulated as:
\begin{equation}
    \vec{n} = \frac{\vec{t}\times\vec{p}}{\|~ \vec{t}\times\vec{p} ~\|},
    \label{eq:swing-axis}
\end{equation}
and the \textit{swing} angle $\alpha$ satisfies:
\begin{equation}
    \cos{\alpha} = \frac{\vec{t}\cdot\vec{p}}{\|\vec{t}\| \|\vec{p}\|}, \quad
    \sin{\alpha} = \frac{\|~ \vec{t}\times\vec{p} ~\|}{\|\vec{t}\| \|\vec{p}\|}.
    \label{eq:swing-angle}
\end{equation}
Hence, the closed-form solution of the \textit{swing} rotation $R^{\textit{sw}}$ can be derived by the \textit{Rodrigues formula}:
\begin{equation}
    R^{\textit{sw}} = \mathcal{D}^{\textit{sw}}(\vec{p}, \vec{t}) = \mathcal{I} + \sin{\alpha}[\vec{n}]_{\times} + (1 - \cos{\alpha})[\vec{n}]_{\times}^2,
    \label{eq:swing-rod}
\end{equation}
where $[\vec{n}]_{\times}$ is the skew symmetric matrix of $\vec{n}$ and $\mathcal{I}$ is the $3\times3$ identity matrix.

\textit{\textbf{- Twist}:}\quad
The \textit{twist} rotation is rotating around $\vec{t}$ itself. Thus, with $\vec{t}$ itself the axis and $\phi$ the angle, we can determine \textit{twist} rotation $R^{tw}$:
\begin{equation}
    R^{\textit{tw}} = \mathcal{D}^{\textit{tw}}(\vec{t}, \phi) = \mathcal{I} + \frac{\sin{\phi}}{\| \vec{t} \|}[\vec{t}]_{\times} + \frac{(1 - \cos{\phi})}{\| \vec{t} \|^2}[\vec{t}]_{\times}^2,
    \label{eq:twist-rod}
\end{equation}
where $[\vec{t}]_{\times}$ is the skew symmetric matrix of $\vec{t}$.

Note that function $\mathcal{D}^{\textit{sw}}$ and $\mathcal{D}^{\textit{tw}}$ are fully differentiable, which allows us to integrate the twist-and-swing decomposition into the training process.
Although we need a neural network to learn the \textit{twist} angle, the difficulty of learning is significantly reduced.
Compared with the 3-DoF rotation that is directly regressed in previous work~\cite{hmr,spin,vibe}, the \textit{twist} angle is only a 1-DoF variable. Moreover, due to the physical limitation of the human body, the \textit{twist} angle has a small range of variation. Therefore, it is much easier for the networks to learn the mapping function. We further analyze its variation in \S\ref{sec:ablation}. 

\vspace{-2mm}
\paragraph{Naive HybrIK.} The IK process can be performed recursively along the kinematic tree like the FK process. First of all, we need to determine the global root rotation $R_{0}$, which has a closed-form solution using the locations of $\mathtt{spine}$, $\mathtt{left~hip}$, $\mathtt{right~hip}$ and Singular Value Decomposition (SVD). Detailed mathematical proof is provided in the supplemental document. Then, in each step, \eg the $k$-th step, we assume the rotation of the parent joint $R_{\mathtt{pa}(k)}$ is known. Hence, we can reformulate Eq.~\ref{eq:ik-cond} with Eq.~\ref{eq:r-step} as:
\begin{equation}
    R_{\mathtt{pa}(k)}^{-1}(p_k - p_{\mathtt{pa}(k)}) = R_{\mathtt{pa}(k), k}(t_k - t_{\mathtt{pa}(k)}).
\end{equation}
Let $\vec{p}_k = R_{\mathtt{pa}(k)}^{-1}(p_k - p_{\mathtt{pa}(k)})$ and $\vec{t}_k = (t_k - t_{\mathtt{pa}(k)})$, we can solve the relative rotation via Eq.~\ref{eq:dec}:
\begin{equation}
    R_{\mathtt{pa}(k), k} = \mathcal{D}(\vec{p}_k, \vec{t}_k, \phi_k),
\end{equation}
where $\phi_k$ is the network predicting $\textit{twist}$ angle for the $k$-th joint. The set of \textit{twist} angle is denoted as $\Phi = \{\phi_k\}_{k=1}^K$. Since the rotation matrices are orthogonal, their inverse equals to their transpose, \ie $R_{\mathtt{pa}(k)}^{-1} =  R_{\mathtt{pa}(k)}^\mathrm{T}$, which keeps the solving process differentiable.

The whole process is named Naive HybrIK and summarized in Alg.~\ref{alg:naive-hybrik}.
Note that we solve the relative rotation $R_{\mathtt{pa}(k), k}$ instead of the global rotation $R_k$.
The reason is that if we directly decompose the global rotation, the resulting \textit{twist} angle will depend on all ancestors' rotations along the kinematic tree, which increases the variation of the distal limb joints and the difficulty for the network to learn.

{
\begin{algorithm}[t]
    \label{alg:naive-hybrik}
    \caption{Naive HybrIK}
    \KwIn{$\mathbf{P}$, $\mathbf{T}$, $\Phi$}
    \KwOut{$\mathbf{R}$}
    Determine $R_{0}$\;
    \For{$k$ along the kinematic tree}
    {
        $\vec{p}_k \leftarrow R_{\mathtt{pa}(k)}^{-1}(p_k - p_{\mathtt{pa}(k)})$\;
        $\vec{t}_k \leftarrow (t_k - t_{\mathtt{pa}(k)})$\;
        $R^{\textit{sw}}_{\mathtt{pa}(k),k} \leftarrow \mathcal{D}^{\textit{sw}}(\vec{p}_k, \vec{t}_k)$\;
        $R^{\textit{tw}}_{\mathtt{pa}(k),k} \leftarrow \mathcal{D}^{\textit{tw}}(\vec{t}_k, \phi_k)$\;
        $R_{\mathtt{pa}(k),k} \leftarrow R^{\textit{sw}}_{\mathtt{pa}(k),k} R^{\textit{tw}}_{\mathtt{pa}(k),k}$\;
    }
\end{algorithm}
}

\paragraph{Adaptive HybrIK.} Although the Naive HybrIK process seems effective, it follows an unstated hypothesis: $\| p_k - p_{\mathtt{pa}(k)} \| = \| t_k - t_{\mathtt{pa}(k)} \|$. Otherwise, there is no solution for Eq.~\ref{eq:ik-cond}. Unfortunately, in our case, the body-parts predicted by the 3D keypoint estimation method are not always consistent with the rest pose template. In Naive HybrIK, Eq.~\ref{eq:dec} can still be solved because the condition is turned into:
\begin{equation}
    p_k - p_{\mathtt{pa}(k)} = R_k(t_k - t_{\mathtt{pa}(k)}) + \vec{\epsilon}_k,
    \label{eq:ik-cond-2}
\end{equation}
where $\vec{\epsilon}_k$ denotes the error in the $k$-th step, which has the same direction of $p_k - p_{\mathtt{pa}(k)}$ and $\| \vec{\epsilon}_k \| = \lvert \| p_k - p_{\mathtt{pa}(k)} \| - \| t_k - t_{\mathtt{pa}(k)} \| \rvert$. 
To analyze the reconstruction error, we compare the difference between the input pose $\mathbf{P}$ and the reconstructed pose $\mathbf{Q}$:
\begin{equation}
    \| \mathbf{P} - \mathbf{Q} \| \Leftrightarrow \sum_{k=1}^K \| p_k - q_k \|,
\end{equation}
where $\mathbf{Q} = \text{FK}(\mathbf{R}, \mathbf{T}) = \text{FK}(\text{IK}(\mathbf{P}, \mathbf{T}), \mathbf{T})$. Combining Eq.~\ref{eq:fk-step} and Eq.~\ref{eq:ik-cond-2}, we have:
\begin{equation}
    \begin{aligned}
    p_k - q_k &= p_{\mathtt{pa}(k)} - q_{\mathtt{pa}(k)} + \vec{\epsilon}_k \\
    &= p_{\mathtt{pa}^2(k)} - q_{\mathtt{pa}^2(k)} + \vec{\epsilon}_{\mathtt{pa}(k)} + \vec{\epsilon}_k \\
    &= \ldots =  \sum_{i \in A(k)} \vec{\epsilon}_i,
    \end{aligned}
    \label{eq:naive-error}
\end{equation}
where $\mathtt{pa}^2(k)$ denotes the parent index of the $\mathtt{pa}(k)$-th joint, and $A(k)$ denotes the set of ancestors of the $k$-th joint. That means the difference between the input joint $p_k$ and the reconstructed joint $q_k$ will accumulate along the kinematic tree, which brings more uncertainty to the distal joint.

\begin{figure}[tb]
    \begin{center}
        \includegraphics[width=0.95\linewidth]{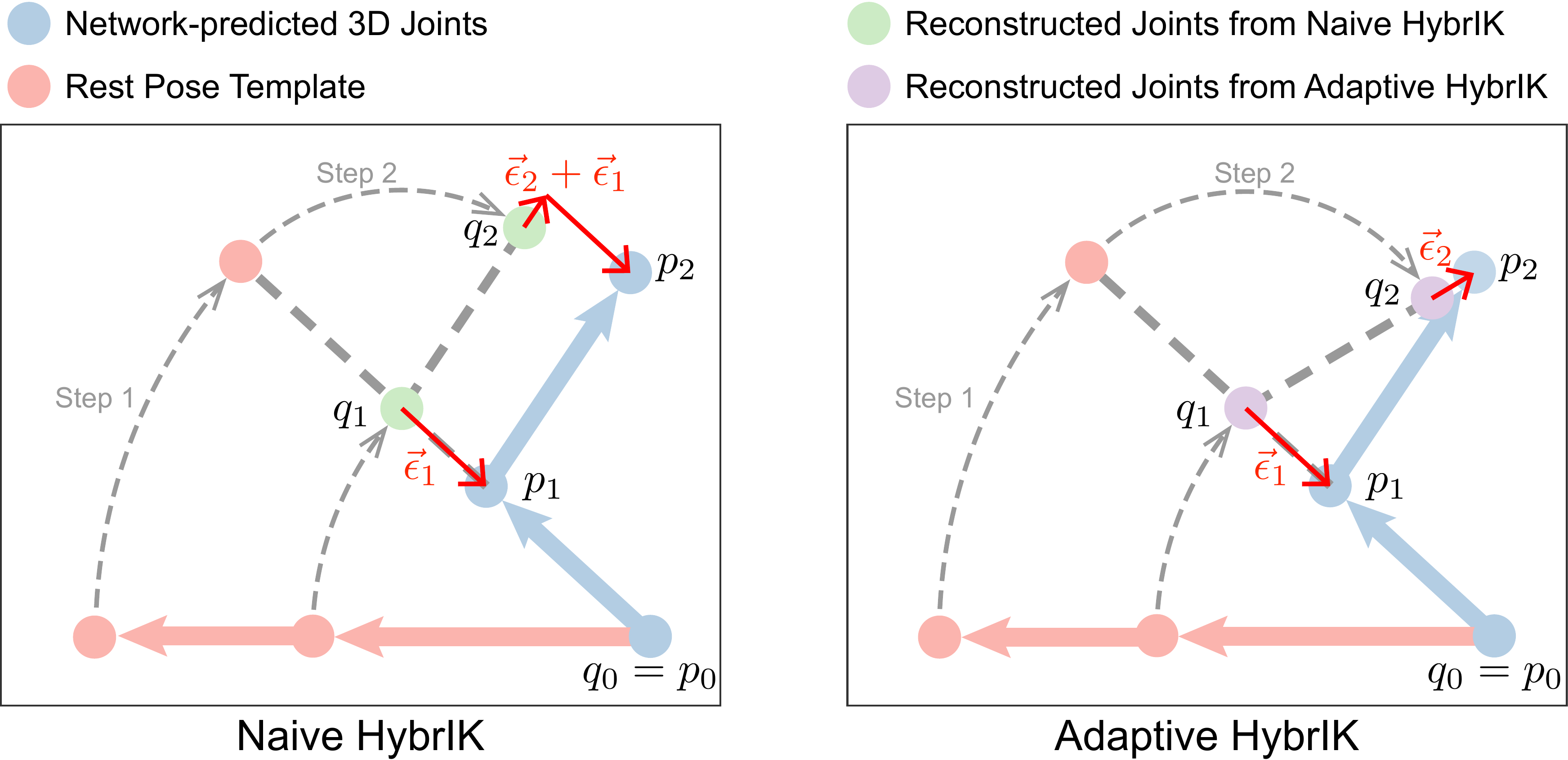}
    \end{center}
    \vspace{-3mm}
    \caption{\textbf{Example of the reconstruction error.} The rest pose is rotated to $q_1$ and $q_2$ by two steps. In the first step, due to the bone-length inconsistency, the reconstruction error is $\vec{\epsilon}_1$. In the second step, Naive HybrIK takes $p_2 - p_1$ as the target direction, resulting in the accumulation of error $\vec{\epsilon}_1 + \vec{\epsilon}_2$. Instead, Adaptive HybrIK selects the reconstructed joint $q_1$ to form the target direction $p_2 - q_1$, which reduces the error to only $\vec{\epsilon}_2$.}
    \label{fig:error}
    \vspace{-3mm}
\end{figure}

To address this error accumulation problem, we further propose Adaptive HybrIK. In Adaptive HybrIK, the target vector is adaptively updated by the newly reconstructed parent joints.
Let $\vec{p}_k = R_{\mathtt{pa}(k)}^{-1}(p_k - q_{\mathtt{pa}(k)})$ and $\vec{t}_k$ the same as the one in the naive solution. In this way, the condition in Adaptive HybrIK can be formulated as:
\begin{equation}
    p_k - q_{\mathtt{pa}(k)} = R_k(t_k - t_{\mathtt{pa}(k)}) + \vec{\epsilon}_k.
    \label{eq:ik-cond-3}
\end{equation}
Therefore, we have:
\begin{equation}
    \begin{aligned}
    &{p_k} - q_{\mathtt{pa}(k)} = q_k - q_{\mathtt{pa}(k)} + \vec{\epsilon}_k \\
    \Rightarrow &\quad p_k - q_k = \vec{\epsilon}_k.
    \end{aligned}
\end{equation}
Compared to the naive solution (Eq.~\ref{eq:naive-error}), the reconstructed error of the adaptive solution only depends on the current joint and will not accumulate from its ancestors. As illustrated in Fig.~\ref{fig:error}, in Naive HybrIK, once the parent joint is out of position, its children will continue this mistake. Instead, in Adaptive HybrIK, the solved relative rotation is always pointing towards the target joint and reduce the error. We conduct empirical experiments in \S\ref{sec:ablation} to validate its robustness.
Note that an iterative global optimization process can further reduce the error, but it is non-differentiable and does not allow end-to-end training.
Adaptive HybrIK is robust enough and remains differentiable. The whole process is summarized in Alg.~\ref{alg:adaptive-hybrik}.

\subsection{Learning Framework}\label{sec:learning}
The overall framework of our approach is illustrated in Fig.~\ref{fig:pipeline}. Firstly, a neural network is utilized to predict 3D joints $\mathbf{P}$, the \textit{twist} angle $\Phi$ and the shape parameters $\beta$. Secondly, the shape parameters are used to obtain the rest pose $\mathbf{T}$ by the SMPL model. Then, by combining $\mathbf{P}$, $\mathbf{T}$ and $\Phi$, we perform HybrIK to solve the relative rotations $\mathbf{R}$ of the 3D pose, \ie the pose parameters $\theta$. Finally, with the function $\mathcal{M}(\theta, \beta)$ provided by the SMPL model, the body mesh $M$ is obtained. The reconstructed pose $\mathbf{Q}$ can be obtained from $M$ by FK or a regressor, which is guaranteed to be realistic. Since the HybrIK process is differentiable, the whole framework is trained in an end-to-end manner.

\vspace{-0.5em}
\paragraph{3D Keypoint Estimation.}
We adopt a simple yet effective architecture to estimate the 3D body keypoints. Following ~\cite{integral}, we use ResNet as our backbone and $3$ deconvolution layers followed by a $1\times1$ convolution to generate the 3D heatmaps. The soft-argmax operation is used to obtain 3D pose from the heatmap in a differentiable manner. We supervise the predicted pose coordinates with $\ell 1$ loss:
\begin{equation}
    \mathcal{L}_{\textit{pose}} = \frac{1}{K} \sum_{k=1}^K \| p_k - \hat{p}_k \|_1,
\end{equation}
where $\hat{p}_k$ denotes the ground-truth joint. 

\begin{algorithm}[t]
    \label{alg:adaptive-hybrik}
    \caption{Adaptive HybrIK}
    \KwIn{$\mathbf{P}$, $\mathbf{T}$, $\Phi$}
    \KwOut{$\mathbf{R}$}
    Determine $R_{0}$\;
    \For{$k$ along the kinematic tree}
    {
        $q_{\mathtt{pa}(k)} \leftarrow R_{\mathtt{pa}(k)}(t_{\mathtt{pa}(k)} - t_{\mathtt{pa}^2(k)}) + q_{\mathtt{pa}^2(k)}$ \;
        $\vec{p}_k \leftarrow R_{\mathtt{pa}(k)}^{-1}(p_k - q_{\mathtt{pa}(k)})$\;
        $\vec{t}_k \leftarrow (t_k - t_{\mathtt{pa}(k)})$\;
        $R^{\textit{sw}}_{\mathtt{pa}(k),k} \leftarrow \mathcal{D}^{\textit{sw}}(\vec{p}_k, \vec{t}_k)$\;
        $R^{\textit{tw}}_{\mathtt{pa}(k),k} \leftarrow \mathcal{D}^{\textit{tw}}(\vec{t}_k, \phi_k)$\;
        $R_{\mathtt{pa}(k),k} \leftarrow R^{\textit{sw}}_{\mathtt{pa}(k),k} R^{\textit{tw}}_{\mathtt{pa}(k),k}$\;
    }
\end{algorithm}

\paragraph{Twist Angle Estimation.}
Instead of the direct regression of scalar value $\phi_k$, we choose to learn a 2-dimensional vector $(c_{\phi_k}, s_{\phi_k})$ that represents $\cos{\phi_k}$ and $\sin{\phi_k}$ to avoid the discontinuity problem.
The $\ell 2$ loss is applied:
\begin{equation}
    \mathcal{L}_{\mathit{tw}} = \frac{1}{K} \sum_{k=1}^K \| (c_{\phi_k}, s_{\phi_k}) -  (\cos{\hat{\phi}_k}, \sin{\hat{\phi}_k}) \|_2,
\end{equation}
where $\hat{\phi}_k$ denotes the ground-truth \textit{twist} angle.

\paragraph{Collaboration with SMPL.} The SMPL model allows us to obtain the rest pose skeleton with the additive offsets according to the shape parameters $\beta$:
\begin{equation}
    \mathbf{T} = W(\bar{M}_{\mathbf{T}} + B_{S}(\beta)),
\end{equation}
where $\bar{M}_{\mathbf{T}}$ is the mesh vertices of mean rest pose, and $B_{S}(\beta)$ is the blend shapes function provided by SMPL.
Then the pose parameters $\theta$ are calculated by HybrIK in a differentiable manner.
In the training phase, we supervise the shape parameters $\beta$:
\begin{equation}
    \mathcal{L}_{\textit{shape}} = \| \beta - \hat{\beta} \|_2,
\end{equation}
and the rotation parameters $\theta$:
\begin{equation}
    \mathcal{L}_{\textit{rot}} = \| \theta - \hat{\theta} \|_2.
\end{equation}
The overall loss of the learning framework is formulated as:
\begin{equation}
    \mathcal{L} = \mathcal{L}_{\textit{pose}} + {\mu_1}\mathcal{L}_{\textit{shape}} + {\mu_2}\mathcal{L}_{\textit{rot}} + {\mu_3}\mathcal{L}_{\textit{tw}},
\end{equation}
where $\mu_1, \mu_2$ and $\mu_3$ are weights of the loss items.

\subsection{Implementation Details}\label{sec:implement}

Here we elaborate more implementation details. We use ResNet-34~\cite{resnet} as the network backbone, initialized with ImageNet pre-trained weights. The ResNet output is divided into two branches. The first branch is to generate 3D heatmaps. The second branch consists of an average pooling, two fully-connected layers with 1024 neurons (each with a dropout layer in between) and a final layer of 56 neurons (10 for $\beta$, 46 for $\Phi$). The input image is resized to $256\times192$. The learning rate is set to $1\times10^{-3}$ at first and reduced by a factor of 10 at the 90th and 120th epoch. We use the Adam solver and train for 140 epochs, with a mini-batch size of $32$ per GPU and 8 GPUs in total. In all experiments, $\mu_1 = 1$ and $\mu_2 = \mu_3 = 1\times10^{-2}$. In the testing phase, the absolute depth of the root joint is obtained from RootNet~\cite{moon2019camera}. Implementation is in PyTorch.

\section{Empirical Evaluation}\label{sec:exp}

\begin{figure}[t]
    \begin{center}
        \includegraphics[width=0.9\linewidth]{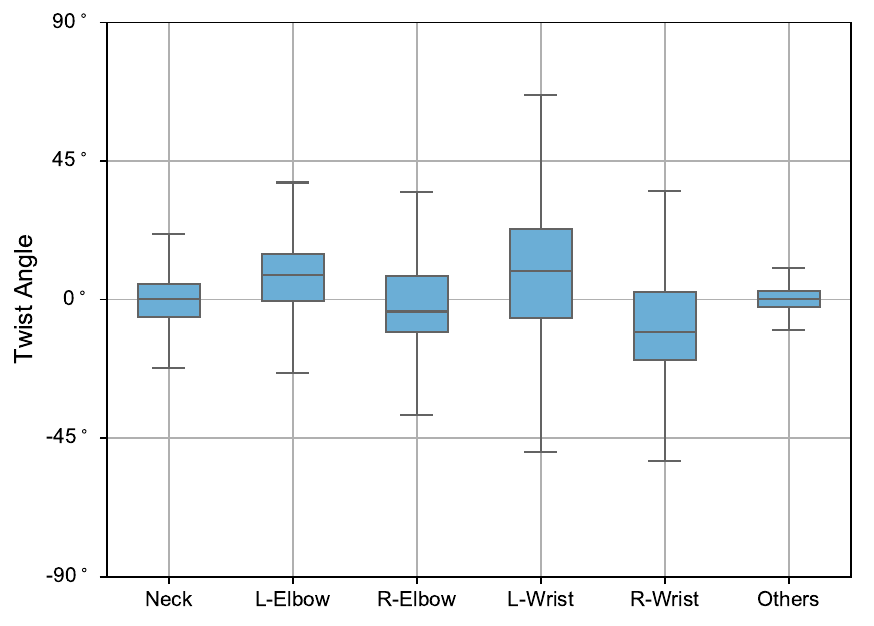}
    \end{center}
    \vspace{-5mm}
    \caption{\textbf{Distribution of the \textit{twist} angle}. Only a few joints have a range over $30^{\circ}$. Other joints have a limited range of \textit{twist} angle.}
    \label{fig:dist-twist}
    \vspace{-2mm}
\end{figure}

In this section, we first describe the datasets employed for training and quantitative evaluation.
Next, ablation experiments are conducted to evaluate the proposed HybrIK. Finally, we report our results and compare the proposed method with state-of-the-art approaches.

\subsection{Datasets}

\noindent\textbf{3DPW:}\quad
It is a challenging outdoor benchmark for 3D pose and shape estimation.
We use this dataset for both training and evaluation.

\noindent\textbf{MPI-INF-3DHP:}\quad
It consists of both constrained indoor and complex outdoor scenes.
Following \cite{hmr,spin}, we use its train set for training and evaluate on its test set.

\noindent\textbf{Human3.6M:}\quad
It is an indoor benchmark for 3D pose estimation.
Following \cite{hmr,spin}, we use 5 subjects (S1, S5, S6, S7, S8) for training and 2 subjects (S9, S11) for evaluation.

\noindent\textbf{MSCOCO:}\quad
It is a large-scale in-the-wild 2D human pose datasets.
We incorporate its train set for training.

\begin{table}[t]
    \begin{center}
    \resizebox{\linewidth}{!}
    {%
        \setlength{\tabcolsep}{1mm}{
        \begin{tabular}{l|ccc|ccc|ccc}
        \toprule
        & \multicolumn{3}{c}{Random Twist} & \multicolumn{3}{c}{Estimated Twist} & \multicolumn{3}{c}{Zero Twist} \\
        \cmidrule(lr){2-4} \cmidrule(lr){5-7} \cmidrule(lr){8-10}
        ~& 24 \textit{jts} & 14 \textit{jts} & Vert. & 24 \textit{jts} & 14 \textit{jts} & Vert. & 24 \textit{jts} & 14 \textit{jts} & Vert.  \\
        \midrule
        Error & 0.1 & 40.0 & 67.3 & 0.1 & 6.1 & 10.0 & 0.1 & 6.8 & 12.1 \\
        \bottomrule
        \end{tabular}
        }
    }
    \end{center}
    \vspace{-2mm}
    \caption{\textbf{Reconstruction error with different \textit{twist} angle.} The accurate \textit{twist} angles significantly reduce the reconstruction error.}
    \label{tab:twist}
    \vspace{-3mm}
\end{table}

\begin{table*}
    \begin{center}
    \resizebox{\textwidth}{!}
    {%
        \begin{tabular}{l|ccc|cc|ccc}
        \toprule
        & \multicolumn{3}{c}{3DPW} & \multicolumn{2}{c}{Human3.6M} & \multicolumn{3}{c}{MPI-INF-3DHP} \\
        \cmidrule(lr){2-4} \cmidrule(lr){5-6} \cmidrule(lr){7-9}
        Method & PA-MPJPE $\downarrow$ & MPJPE $\downarrow$ & PVE $\downarrow$ & PA-MPJPE $\downarrow$ & MPJPE $\downarrow$ & PCK $\uparrow$ & AUC $\uparrow$ & MPJPE $\downarrow$ \\
        \midrule
        \cellcolor{White}SMPLify~\cite{bogo2016keep} & - & \cellcolor{White}- & - & \cellcolor{White}82.3 & - & \cellcolor{White}- & \cellcolor{White}- & \cellcolor{White}- \\
        \cellcolor{Gray}HMR~\cite{hmr} & \cellcolor{Gray}81.3 & \cellcolor{Gray}130.0 & \cellcolor{Gray}- & \cellcolor{Gray}56.8 & \cellcolor{Gray}88.0 & \cellcolor{Gray}72.9 & \cellcolor{Gray}36.5 & \cellcolor{Gray}124.2 \\
        \cellcolor{White}Kolotouros \etal~\cite{cmr} & 70.2 & \cellcolor{White}- & - & \cellcolor{White}50.1 & - & \cellcolor{White}- & \cellcolor{White}- & \cellcolor{White}- \\
        \cellcolor{Gray}Pavlakos \etal~\cite{pavlakos2018learning} & \cellcolor{Gray}- & \cellcolor{Gray}- & \cellcolor{Gray}- & \cellcolor{Gray}75.9 & \cellcolor{Gray}- & \cellcolor{Gray}- & \cellcolor{Gray}- & \cellcolor{Gray}- \\
        \cellcolor{White}Arnab \etal~\cite{arnab2019exploiting} & \cellcolor{White}72.2 & - & - & \cellcolor{White}54.3 & 77.8 & \cellcolor{White}- & \cellcolor{White}- & \cellcolor{White}- \\
        \cellcolor{Gray}SPIN~\cite{spin} & \cellcolor{Gray}59.2 & \cellcolor{Gray}96.9 & \cellcolor{Gray}116.4 & \cellcolor{Gray}41.1 & \cellcolor{Gray}- & \cellcolor{Gray}76.4 & \cellcolor{Gray}37.1 & \cellcolor{Gray}105.2 \\
        \cellcolor{White}I2L~\cite{i2l}$^*$ & \cellcolor{White}58.6 & 93.2 & - & \cellcolor{White}41.7 & 55.7 & \cellcolor{White}- & \cellcolor{White}- & \cellcolor{White}- \\
        \cellcolor{Gray}Mesh Graphormer~\cite{lin2021mesh} \textit{w. 3DPW} & \cellcolor{Gray}45.6 & \cellcolor{Gray}74.7 & \cellcolor{Gray}87.7 & \cellcolor{Gray}41.2 & \cellcolor{Gray}34.5 & \cellcolor{Gray}- & \cellcolor{Gray}- & \cellcolor{Gray}- \\
        \cellcolor{White}PARE~\cite{kocabas2021pare} \textit{w. 3DPW} & \cellcolor{White}46.4 & 74.7 & 87.7 & \cellcolor{White}- & - & \cellcolor{White}- & \cellcolor{White}- & \cellcolor{White}- \\
        \midrule
        \cellcolor{Gray}Ours (Naive HybrIK) & \cellcolor{Gray}\text{49.0} & \cellcolor{Gray}\text{80.2} & \cellcolor{Gray}\text{94.6} & \cellcolor{Gray}35.3 & \cellcolor{Gray}55.8 & \cellcolor{Gray}\text{85.9} & \cellcolor{Gray}\text{41.7} & \cellcolor{Gray}\text{91.5} \\
        Ours (Adaptive HybrIK) & {48.8} & {80.0} & {94.5} & {34.5} & \textbf{54.4} & {86.2} & {42.2} & \textbf{91.0} \\
        Ours (Adaptive HybrIK) \textit{w. 3DPW} & \textbf{45.0} & \textbf{74.1} & \textbf{86.5} & \textbf{33.6} & {55.4} & \textbf{87.5} & \textbf{46.9} & {93.9} \\
        \bottomrule
        \end{tabular}
    }
    \end{center}
    \vspace{-3mm}
    \caption{\textbf{Benchmark of state-of-the-art models on 3DPW, Human3.6M and MPI-INF-3DHP datasets.} ``$*$'' denotes the method is trained on different datasets. ``-'' shows the results that are not available.}
    \label{tab:body}
    \vspace{-3mm}
\end{table*}

\subsection{Ablation Study}\label{sec:ablation}
In this study, we evaluate the effectiveness of the twist-and-swing decomposition and the HybrIK algorithm. All evaluation is conducted on the 3DPW test set as it contains challenging in-the-wild scenes to demonstrate the strength of our model. More experimental results are provided in the supplemental document.

\vspace{-0.7em}
\paragraph{Analysis of the twist rotation.} To demonstrate the effectiveness of twist-and-swing decomposition, we first count the distribution of the \textit{twist} angle in the 3DPW test set. The distribution is illustrated in Fig.~\ref{fig:dist-twist}. As expected, due to the physical limitation, only $\mathtt{neck}$, $\mathtt{elbow}$ and $\mathtt{wrist}$ have a wide range of variations. All other joints have a limited range of \textit{twist} angle (less than 30$^\circ$).
It indicates that the \textit{twist} angle can be reliably estimated.

Besides, we develop an experiment to see how the \textit{twist} angles affect the reconstructed pose and shape. We take the ground-truth 24 SMPL joints and shape parameters as the input of the HybrIK process. As for the \textit{twist} angle, we compare random values in $[-{\pi}, {\pi}]$ and the values estimated by the network. We evaluation the mean error of the reconstructed 24 SMPL joints, the 14 LSP joints, the body mesh and the \textit{twist} angle. Here, following previous works~\cite{bogo2016keep,hmr,spin}, the 14 LSP joints are regressed from the body mesh by a pretrained regressor. Quantitative results are reported in Tab.~\ref{tab:twist}. It shows that the accurate regressed \textit{twist} angles significantly reduce the error on the mesh vertices and the LSP joints that regressed from the mesh. Since most of the twist angles are close to zeros, the zero twist angles produce acceptable performance. Notice that the wrong \textit{twist} angles do not affect the reconstructed SMPL joints. Only the \textit{swing} rotations change the joint locations.


\begin{table}
    \begin{center}
    \resizebox{\linewidth}{!}
    {%
        \begin{tabular}{l|c|c|c|c}
            \toprule
            ~& GT Joints & $\pm 10$ mm & $\pm 20$ mm & $\pm 30$ mm \\
            \midrule
            Naive HybrIK & 0.1 & 16.2 & 34.0 & 53.4 \\
            Adaptive HybrIK & 0.1 & 9.8 & 20.2 & 31.2 \\
            \bottomrule
        \end{tabular}
    }
    \end{center}
    \vspace{-3mm}
    \caption{\textbf{Naive \textit{vs.} Adaptive} with different input joints. MPJPE of 24 joints is reported. Adaptive HybrIK is more robust to the jitters.}
    \label{tab:robust}
    \vspace{-3mm}
\end{table}

\vspace{-0.7em}
\paragraph{Robustness of HybrIK.}
To demonstrate the superiority of Adaptive HybrIK over Naive HybrIK, we compare the reconstructed joints error of these two algorithms. First, we feed the ground-truth joints, \textit{twist} angle and shape parameters to the two IK algorithms to see whether they will introduce extra error. Then we add jitters to the input to observe the performance of the HybrIK algorithm with the noisy joints.
As shown in Tab.~\ref{tab:robust}, when the input joints are correct, both HybrIK algorithms introduce negligible errors.
For noisy joints input, the Naive HybrIK algorithm accumulates errors along the kinematic tree, while the Adaptive HybrIK algorithm is more robust to the noise.

\paragraph{Error correction capability of HybrIK.}
In this experiment, we examine the error correction capability of the HybrIK algorithm. The HybrIK algorithm is fed with the 3D joints, \textit{twist} angles and shape parameters that predicted by the neural network. Additionally, we apply the SMPLify~\cite{bogo2016keep} algorithm on the predicted pose and compare it to our method. As shown in Tab.~\ref{tab:pat-ik}, 
the error of reconstructed joints after HybrIK is reduced the error to $79.2$mm, while SMPLify raises the error to 114.3 mm.
The error correction capability of HybrIK comes from the fact that the network may predict unrealistic body pose, \eg left-right asymmetry and abnormal limbs proportions. In contrast, the rest pose is generated by the parametric statistical body model, which guarantees that the reconstructed pose is consistent with the realistic body shape distribution.
Since our proposed framework is agnostic to the way we obtain 3D joints, we can improve the performance of any single-stage 3D keypoint estimation methods.

\subsection{Comparison with the State-of-the-art}
To make a fair comparison with previous 3D human pose and shape estimation methods, we use a regressor to obtain the $14$ LSP joints from the body mesh for the evaluation on 3DPW and Human3.6M datasets and $17$ joints for MPI-INF-3DHP dataset.
Procrustes aligned mean per joint position error (PA-MPJPE), mean per joint position error (MPJPE), Percentage of Correct Keypoints (PCK) and Area Under Curve (AUC) are reported to evaluate the 3D pose results. We also report Per Vertex Error (PVE) to evaluate the entire estimated body mesh.

In Tab.~\ref{tab:body}, we compare our method with previous 3D human pose and shape estimation methods, including both model-based and model-free methods, on 3DPW, Human3.6M and MPI-INF-3DHP datasets.
Without bells and whistles, our method surpasses all previous state-of-the-art methods by a large margin on all three datasets. It is worth noting that our method improve $21.9$ mm PVE on 3DPW dataset, which shows that it is accurate and reliable to recover body mesh through inverse kinematics.

\begin{table}
    \begin{center}
    \resizebox{\linewidth}{!}
    {%
        \begin{tabular}{l|c|c|c}
        \toprule
        ~& Predicted Pose & HybrIK & SMPLify~\cite{bogo2016keep} \\
        \midrule
        MPJPE (24 \textit{jts}) $\downarrow$ & 88.2 mm & 79.2 mm & 114.3 mm \\
        \bottomrule
        \end{tabular}
    }
    \end{center}
    \vspace{-3mm}
    \caption{\textbf{Error correction capability of HybrIK}. HybrIK improves the results predicted by the 3D keypoint estimation method.}
    \label{tab:pat-ik}
    \vspace{-3mm}
\end{table}




\section{Conclusion}
In this paper, we bridge the gap between 3D keypoint estimation and body mesh estimation via a novel hybrid analytical-neural inverse kinematics solution, HybrIK. It transforms the 3D joint locations to a pixel-aligned accurate human body mesh, and then obtains a more accurate and realistic 3D skeleton from the reconstructed 3D mesh, closing the loop between the 3D skeleton and the parametric body model. Our method is fully differentiable and allows simultaneously training of 3D joints and human body mesh in an end-to-end manner. We demonstrate the effectiveness of our method on various 3D pose and shape datasets. The proposed method surpasses state-of-the-art methods by a large margin. Besides, comprehensive analyses demonstrate that HybrIK is robust and has error correction capability. We hope HybrIK can serve as a solid baseline and provide a new perspective for the 3D human pose and shape estimation task.

\paragraph{Acknowledgements}
This work is supported in part by the National Key R\&D Program of China, No. 2017YFA0700800, National Natural Science Foundation of China under Grants 61772332 and Shanghai Qi Zhi Institute, SHEITC (018-RGZN-02046).

{\small
\bibliographystyle{ieee_fullname}
\bibliography{egbib}
}

\newpage
\section*{Appendix}
\label{sec:appendix}

\subsection*{A \quad Rigid Registration of Global Rotation}
\label{sec:app-a}

In the SMPL model~\cite{loper2015smpl}, the pose parameters $\theta$ control the rotations of the rigid body parts. The three joints named $\mathtt{spine}$, $\mathtt{left~hip}$ and $\mathtt{right~hip}$ form a rigid body part, which is controlled by the global root rotation. Therefore, the global rotation can be determined by registering the rest pose template of $\mathtt{spine}$, $\mathtt{left~hip}$ and $\mathtt{right~hip}$ to the predicted locations of these three joints. Let $t_1$, $t_2$ and $t_3$ denote their locations in the rest pose template, and $p_1$, $p_2$ and $p_3$ denote the predicted locations. Our goal is to find a rigid rotation that optimally aligns the two sets of joints. Here, we assume the root joint of the predicted pose and the rest pose are aligned. Hence, the problem is formulated as:
\begin{equation}
   R_0 = \arg\min_{R \in \mathbb{SO}^3} \sum_{i=1}^3 \| p_i - R t_i \|_2^2.
\end{equation}
This formula can be written in matrix form:
\begin{equation}
   R_0 = \arg\min_{R \in \mathbb{SO}^3} \| P_0 - R T_0 \|_F^2,
   \label{eq:problem}
\end{equation}
where $\| \cdot \|_F$ denotes the Frobenius norm,$P_0$ denotes $[p_0~p_1~p_2]$, and $T_0$ denotes $[t_0~t_1~t_2]$. Let us simplify the expression in Eq.~\ref{eq:problem} as:
\begin{equation}
   \begin{aligned}
      &\min_{R \in \mathbb{SO}^3} \| P_0 - R T_0 \|_F^2 \\
      \Leftrightarrow &\min_{R \in \mathbb{SO}^3} \text{trace}((P_0 - R T_0)^\mathrm{T}(P_0 - R T_0)) \\
      \Leftrightarrow &\min_{R \in \mathbb{SO}^3} \text{trace}(P_0^\mathrm{T}P_0 + T_0^\mathrm{T}T_0 - 2P_0^\mathrm{T}RT_0). \\
   \end{aligned}
\end{equation}
Note that $P_0^\mathrm{T}P_0$ and $T_0^\mathrm{T}T_0$ are independent of $R$. Thus the original problem is equivalent to:
\begin{equation}
   \begin{aligned}
   &\arg\min_{R \in \mathbb{SO}^3} \| P_0 - R T_0 \|_F^2 \\
   \Leftrightarrow &\arg\max_{R \in \mathbb{SO}^3} \text{trace}(P_0^\mathrm{T}RT_0).
   \end{aligned}
\end{equation}
Further, we can leverage the property of the matrix trace,\begin{equation}
   \text{trace}(P_0^\mathrm{T}RT_0) = \text{trace}(RT_0P_0^\mathrm{T}).
\end{equation}
Then, we apply Singular Value Decomposition (SVD) to the joint locations:
\begin{equation}
   T_0P_0^\mathrm{T} = U \Lambda V^\mathrm{T}.
\end{equation}
The problem is equivalent to:
\begin{equation}
   \begin{aligned}
   &\arg\max_{R \in \mathbb{SO}^3} \text{trace}(RT_0P_0^\mathrm{T}) \\
   \Leftrightarrow &\arg\max_{R \in \mathbb{SO}^3} \text{trace}(R U \Lambda V^\mathrm{T}) \\
   \Leftrightarrow &\arg\max_{R \in \mathbb{SO}^3} \text{trace}(\Lambda V^\mathrm{T}R U).
   \end{aligned}
\end{equation}
Note that $U$, $V$ and $R$ are orthogonal matrices, so $M = V^\mathrm{T}RU$ is also an orthogonal matrix. Then, for all $1 \leq j \leq$ we have:
\begin{equation}
   \begin{aligned}
   &m_j^\mathrm{T}m_j = 1 = \sum_{i=1}^3 m_{ij}^2 \\
   \Rightarrow ~ &m_{ij}^2 \leq 1 \Rightarrow | m_{ij} | \leq 1.
   \end{aligned}
\end{equation}
Besides, $\Lambda$ is a diagonal matrix
with non-negative values, \ie $\lambda_1$, $\lambda_2$, $\lambda_3 \geq 0$. Therefore:
\begin{equation}
   \begin{aligned}
   \text{trace}(\Lambda V^\mathrm{T}R U) &= \text{trace}(\Lambda M) \\ &= \sum_{i=1}^3 \lambda_i m_{ii}
   \leq \sum_{i=1}^3 \lambda_i.
   \end{aligned}
\end{equation}
The trace is maximized if $m_{ii} = 1, \forall 1 \leq i \leq 3$. That means $M = \mathcal{I}$, where $\mathcal{I}$ is the identity matrix. Finally, the optimal rotation $R_0$ is:
\begin{equation}
   \begin{aligned}
   V^\mathrm{T}R_0 U = \mathcal{I} \\
   \Rightarrow
   R_0 = V U^\mathrm{T}.
   \end{aligned}
\end{equation}

\section*{B. \quad More Ablation Experiments}

\begin{table}[h]
   \begin{center}
   \resizebox{\linewidth}{!}
   {%
         \begin{tabular}{l|cc|cc|cc}
         \toprule
         & \multicolumn{2}{c}{GT $\beta$} & \multicolumn{2}{c}{Estimated $\beta$} & \multicolumn{2}{c}{Zero $\beta$} \\
         \cmidrule(lr){2-3} \cmidrule(lr){4-5} \cmidrule(lr){6-7}
         ~& MPJPE & PVE & MPJPE & PVE & MPJPE & PVE  \\
         \midrule
         Error & 72.7 & 87.4 & 80.0 & 94.5 & 81.1 & 95.4 \\
         \bottomrule
         \end{tabular}
   }
   \end{center}
   \caption{\textbf{Reconstruction error with different shape parameters $\beta$.}}
   \label{tab:beta}
\end{table}

\paragraph{Effect of $\beta$}
In this experiment, we analyze the effect of the shape parameters $\beta$ in Tab.~\ref{tab:beta}. Using the ground-truth $\beta$ brings $5$ mm improvement of MPJPE and PVE on 3DPW dataset. Using zero $\beta$ brings $1$ mm error. It shows that there are lots of room for improvement by estimating more accurate $\beta$.

\paragraph{Comparison with Baseline Models}
In this experiment, we compare HybrIK with two baselines to validate its effectiveness. Firstly, we want th compare with the model that directly predicts SMPL parameters without any auxiliary loss. This model is a degraded version of HMR~\cite{hmr}. We find it is hard to train and the model learns limited information. The model achieves over $100$ mm error on Human3.6M~\cite{h36m}. Secondly, we add 3D keypoint prediction to help the network to extract features. The model still learns to predict SMPL parameters directly. However, still over $100$ mm error achieves on Human3.6M~\cite{h36m} dataset.

\begin{table}[t]
   \begin{center}
   \resizebox{\linewidth}{!}
   {%
         \begin{tabular}{l|cc|cc}
         \toprule
         & \multicolumn{2}{c}{Human3.6M} & \multicolumn{2}{c}{3DPW} \\
         \cmidrule(lr){2-3} \cmidrule(lr){4-5}
         ~& Predicted Pose & HybrIK & Predicted Pose & HybrIK \\
         \midrule
         MPJPE (24 \textit{jts}) $\downarrow$ & 51.3 & 48.1 & 88.2 & 79.2 \\
         \bottomrule
         \end{tabular}
   }
   \end{center}
   \caption{\textbf{Error correction capability of HybrIK} on 3DPW and Human3.6M.}
   \label{tab:correction}
\end{table}

\paragraph{Error correction capability of HybrIK}
In this experiment, we examine the error correction capability of HybrIK on 3DPW~\cite{3dpw} and Human3.6M~\cite{h36m} datasets. Quantitative results are reported in Tab.~\ref{tab:correction}.

\subsection*{C \quad Qualitative Results}

\begin{figure*}[!ht]
    \centering
    \includegraphics[width=0.9\linewidth]{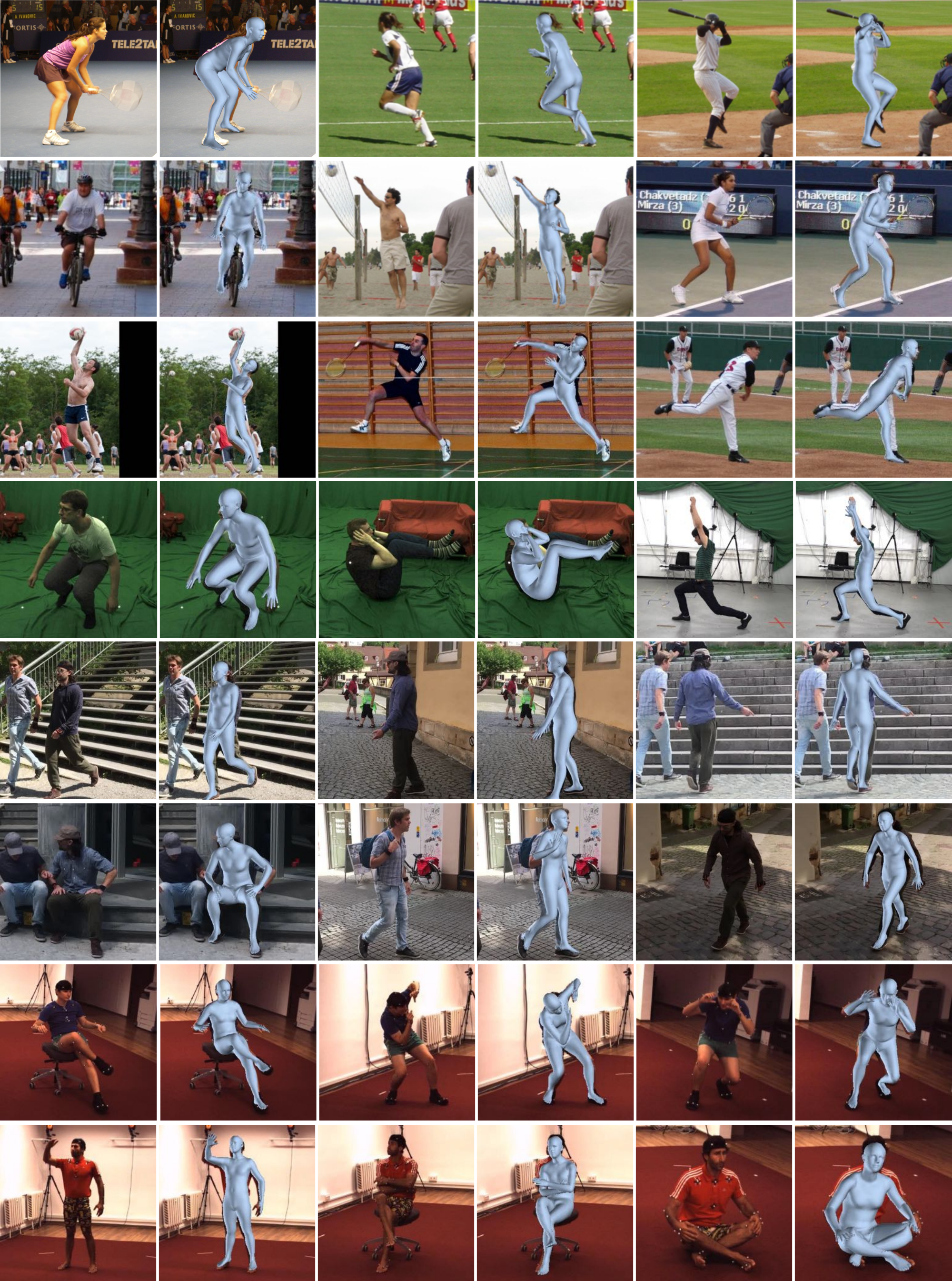}\\
    \caption{Qualitative results from various datasets, LSP (rows 1-3), MPI-INF-3DHP (row 4), 3DPW (rows 5-6), H36M (rows 7-8).}
    \label{fig:qualitative}
\end{figure*}

\begin{figure*}[ht]
    \centering
    \includegraphics[width=0.9\linewidth]{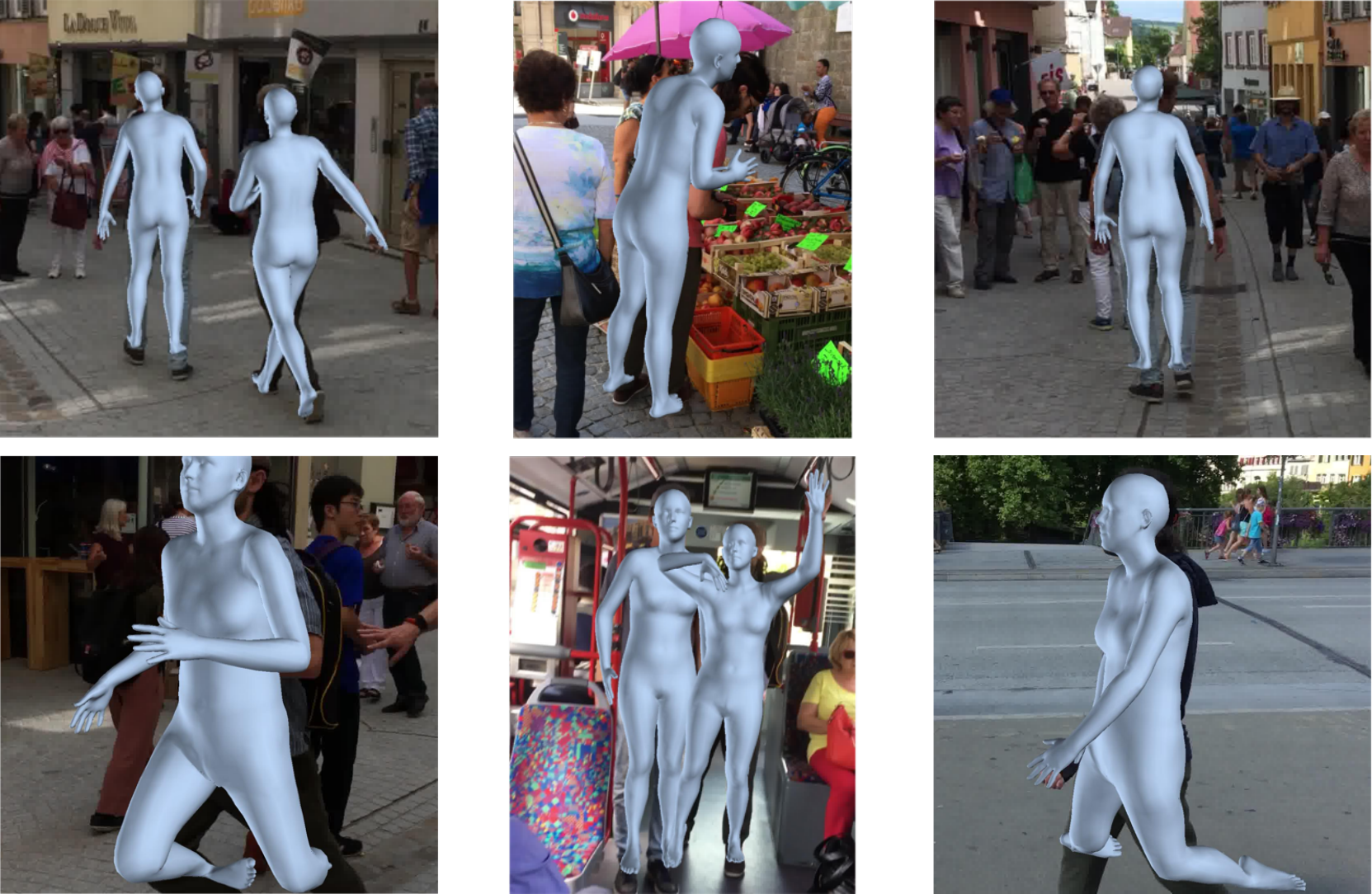}
    \caption{Erroneous reconstructions of our method. Typical failure cases can be attributed to inaccurate bone length estimation (shape parameters $\beta$) and 3D keypoint estimation.}
    \label{fig:erroneous}
\end{figure*}

Fig.~\ref{fig:qualitative} provides qualitative results of our approach from the different datasets involved in our experiments (LSP, MPI-INF-3DHP, Human3.6M, 3DPW). Fig.~\ref{fig:erroneous} includes typical failure cases that are attributed to erroneous bone length estimation (shape parameters $\beta$) and 3D keypoint estimation, which lead to misalignment and unnatural joint bending, respectively.

\end{document}